# CRRN: Multi-Scale Guided Concurrent Reflection Removal Network

Renjie Wan[†,§,∗]  Boxin Shi[★,∗]  Ling-Yu Duan[★]  Ah-Hwee Tan[¶]  Alex C. Kot[§]

[†]Interdisciplinary Graduate School, [§]School of Electrical and Electronic Engineering,
[¶]School of Computer Science and Engineering, Nanyang Technological University, Singapore
[★]National Engineering Laboratory for Video Technology, School of EECS, Peking University, China
rjwan001@e.ntu.edu.sg, {shiboxin,lingyu}@pku.edu.cn, {asahtan,eackot}@ntu.edu.sg

## Abstract

*Removing the undesired reflections from images taken through the glass is of broad application to various computer vision tasks. Non-learning based methods utilize different handcrafted priors such as the separable sparse gradients caused by different levels of blurs, which often fail due to their limited description capability to the properties of real-world reflections. In this paper, we propose the Concurrent Reflection Removal Network (CRRN) to tackle this problem in a unified framework. Our proposed network integrates image appearance information and multi-scale gradient information with human perception inspired loss function, and is trained on a new dataset with 3250 reflection images taken under diverse real-world scenes. Extensive experiments on a public benchmark dataset show that the proposed method performs favorably against state-of-the-art methods.*

## 1. Introduction

Reflections observed in front of the glass significantly degrade the visibility of the scene behind the glass. By obstructing, deforming or blurring the background scene, reflections cause many computer vision systems likely to fail. Reflection removal aims at enhancing the visibility of the background scene while removing the reflections.

Most of the existing reflection removal methods are non-learning based. They rely on the handcrafted image features observed with special assumptions, *e.g.*, the gradient priors on the basis of the different blur levels between background and reflection [27, 16, 14], the ghosting effects [21], the content prior based on the non-local information [28], and so on. However, these specific assumptions are often violated in real-world scenarios, since the low-level image priors they adopt only describe a limited range of the reflection properties and project the partial observation as the whole truth. When the structures and patterns of the background are similar to those of the reflections, the non-learning based methods have difficulty in simultaneously removing reflections and recovering the background [8].

To capture the reflection properties more comprehensively, recent methods have adopted the deep learning to solve this problem [7, 3]. Existing deep learning based methods [7, 3] show improved modeling ability that captures a variety of reflection image characteristics [22, 8]. However, they still adopt a two-stage framework for gradient inference and image inference as many non-learning based methods [27, 14], which do not fully explore the multi-scale information for background recovery. Moreover, they mainly rely on the pixel-wise loss ( $\mathcal{L}_2$ and $\mathcal{L}_1$ loss), that may generate blurry predictions [12]. Last but not least, existing methods are mainly trained with synthetic images which can never capture the comprehensive information in real world image formation process completely.

To address these drawbacks, we propose the Concurrent Reflection Removal Network (CRRN) to remove reflections observed in the wild scenes, as illustrated in Figure 1. Our major contributions are summarized as follows:

- In contrast to the conventional two-stage framework that classifies the gradients, and then recovers the background [27, 4, 14, 23], we combine the two separate stages (gradient inference and image inference) in one unified mechanism to remove reflections concurrently.

- We propose a multi-scale guided learning network to better preserve the background details, where the back-

[∗]Corresponding authors. This research was carried out at the Rapid-Rich Object Search (ROSE) Lab, Nanyang Technological University, Singapore. The ROSE Lab is supported by the National Research Foundation, Singapore, and the Infocomm Media Development Authority, Singapore. This work is supported in part by the Recruitment Program of Global Experts (Youth Program) in China (a.k.a. 1000 Youth Talents), the National Natural Science Foundation of China (61661146005, U1611461, 61390515), the National Key Research and Development Program of China (No.2016YFB1001501), and the NTU-PKU Joint Research Institute through the Ng Teng Fong Charitable Foundation.

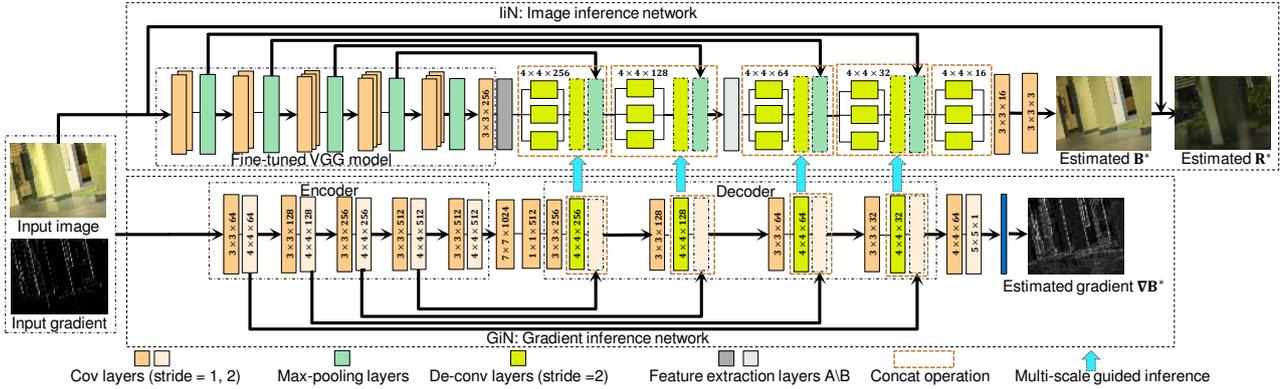

Figure 1. The framework of CRRN. It consists of two cooperative sub-networks: the gradient inference network (GiN) to estimate the gradients of the background and the image inference network (IiN) to estimate the background and reflection layers. We feed GiN with the mixture image and its corresponding gradient as a 4-channel tensor and IiN with the mixture image containing reflections. The upsampling stage of IiN is closely guided by the associate gradient features from GiN with the same resolution. IiN consists of two feature extraction layers to extract the scale invariant features related with the background. IiN gives the estimated background and reflection images, while GiN gives the estimated gradient of background as output.

ground reconstruction in the image inference network is closely guided by the associated gradient features in the gradient inference network.

- We design a perceptually motivated loss function, which help suppress the blurry artifacts introduced by the pixel-wise loss functions, and generate better results.

- To facilitate the training of CRNN for general compatibility on real data, we capture a large-scale reflection image dataset to generate training data, which has proved to improve the performance and generality of our method. To the best of our knowledge, this is the first reflection image dataset for data-driven methods.

## 2. Related Work

Reflection removal has been widely studied for more than decades. Previous work can be roughly classified into two categories. The first category solves this problem using the non-learning based methods. Due to the ill-posed nature of this problem, different priors are employed to exploit the special properties of the background and reflection layers. For example, Levin *et al.* [14] adopted the sparsity priors to decompose the input image. However, their method relies on the users to label the background and reflection edges, which is quite labor-intensive and may fail in textured regions. Li *et al.* [16] made use of the different blur levels of the background and reflection layers. Nikolaos *et al.* [1] adopted the Laplacian data fidelity term to solve this problem. Shih *et al.* [?] used the GMM patch prior to remove reflections with the visible ghosting effects. The handcrafted priors adopted by these methods are based on the observations of some special properties between the background and reflection (*e.g.*, different blur levels [27, 16]) which is often violated in the general scenes especially when these properties are weakly observed. Some other methods in this category remove reflections by using a set of images taken from different viewpoints [30, 9]. By exploiting the motion cues between the background and reflection from multiview captures and assuming the glass is closer to the camera, the projected motion of the two layers is different due to the visual parallax. The motion of each layer can be represented by using parametric model, such as the translative motion [2], the affine transformation [9], and the homography [9]. Through the combination of the motion and traditional cues, the non-learning based methods using the multiple images as the input can show more reliable results when the input data are appropriately prepared. However, the requirement for special facilities of capturing limits such methods for practical use, especially for mobile devices or images downloaded from the Internet.

Another category solves the problem by using the learning based methods. Since the deep learning has achieved promising results in both high-level and low-level computer vision problems, its comprehensive modeling ability also benefit reflection removal problems. For example, Paramanand *et al.* [3] proposed a two-stage deep learning approach to learn the edge features of the reflections by using the light field camera. The framework introduced by Fan *et al.* [7] exploited the edge information when training the whole network to better preserve the image details. Though the deep learning based methods can better capture the image properties, the conventional two-stage framework they adopt as many non-learning based methods [27, 14, 13] ignores the intrinsic correlations, which also degrades their performances.

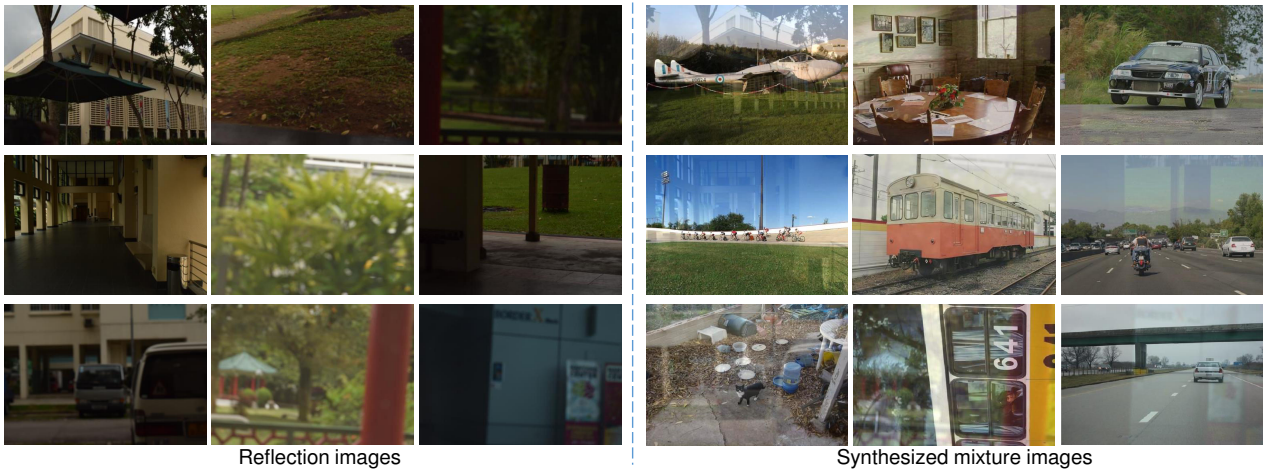

Figure 2. Samples of captured reflection images in the 'RID' and the corresponding synthetic images using the 'RID'. From top to bottom rows, we show the diversity of different illumination conditions, focal lengths, and scenes.

## 3. Dataset Preparation

### 3.1. Real-world refection image dataset for data-driven methods

Real-world image datasets play important roles in studying physics-based computer vision [20] and face anti-spoofing [15] problems. Although the reflection removal problem has been studied for more than decades, publicly available datasets are rather limited. The data-driven methods need a large-scale dataset to learn the reflection image properties. As far as we know, 'SIR$^2$' [26] is the largest reflection removal image datasets, which provides approximately 500 image triplets composed of mixture, background, and reflection images, but its scale is still not sufficient for training a complicated neural network. Considering the difficulty in obtaining the image triplet like 'SIR$^2$', an alternative solution to the data size bottleneck is to use the synthetic image dataset. The recent deep learning based method [7] provides a reasonable way to generate the reflection images by taking the regional properties and blurring effects of the reflections into consideration to make their data similar to the images taken in the wild. However, the ignorance of other reflection image properties (*e.g.*, ghosting effects, various types of noise in the imaging pipeline) may degrade the training and thus limits its wide applicability to real-world scenes.

To facilitate the training of CRRN for general compatibility on real data, we have constructed a large-scale **R**eflection **I**mage **D**ataset called '**RID**', which contains 3250 images in total. We can then use the captured reflection images from the 'RID' to synthesize the input mixture images.

To collect reflection images, we use a NIKON D5300 camera configured with varying exposure parameters and aperture sizes under a fully manual mode to capture images in different scenes. The reflection images are taken by putting a black piece of paper behind the glass while moving the camera and the glass around, which is similar to what have been done in [26, 31].

The 'RID' has the following two major characteristics, with example scenes demonstrated in Figure 2:

- **Diversity.** We consider three aspects to enrich the diversity of the 'RID': 1) We take the reflection images at different illumination conditions to include both strong and weak reflections (the first row in Figure 2 left); 2) we adjust the focal lengths randomly to create different blur levels of reflection. (the second row in Figure 2 left); 3) the reflection images are taken from a great diversity of both indoor and outdoor scenes, *e.g.*, streets, parks, inside of office buildings, and so on (the third row in Figure 2 left).

- **Scale.** The 'RID' has 3250 images in total with approximately 2000 reflection images from the bright scenes and other reflection images are from the relatively dark scenes to meet the request of data-driven methods.

### 3.2. Generating training data

The commonly used image formation model for reflection removal is expressed as:

$$\mathbf{I} = \alpha \mathbf{B} + \beta \mathbf{R}, \tag{1}$$

where $\mathbf{I}$ is the mixture image, $\mathbf{B}$ is the background to be recovered, and $\mathbf{R}$ is the reflection to be removed. In Equation (1), the mixture image $\mathbf{I}$ is a linearly weighted additive of the background $\mathbf{B}$ and the reflection $\mathbf{R}$. Our partially synthetic and partially real training image $\mathbf{I}$ is generated by adding the refection images from the 'RID' as reflection **R** and other natural images (*e.g.*, we use the COCO dataset [17] and the PASCAL VOC dataset [6]) as the background **B** with different weighting factors.

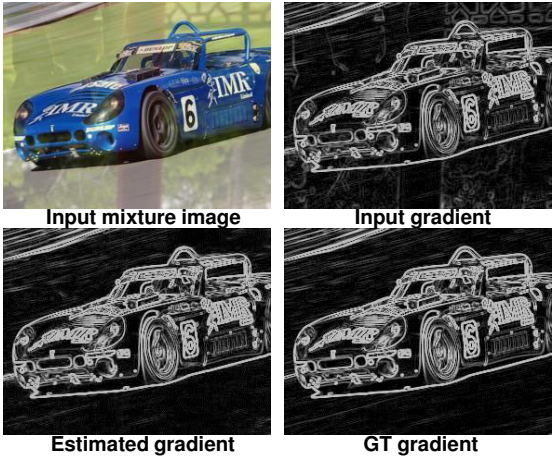

Figure 3. The estimated gradient generated by the gradient inference network, compared with the GT gradient.

To ensure a sufficient amount of training data, $\alpha$ and $\beta$ are randomly sampled from $0.8$ to $1$ and $0.1$ to $0.5$, respectively, and we further augment the generated image with two different operations: image rotation and flipping. In total, our training dataset includes 14754 images.

## 4. Proposed Method

In this section, we describe the design methodology of the proposed reflection removal network, the optimization using human perception inspired loss function, and the details for network training.

### 4.1. Network architecture

According to Equation (1), given the observed images with reflections **I**, our task here is to estimate **B**. Since the estimation of **B** and **R** are intrinsically correlated and the gradient information $\nabla \mathbf{B}$ has been proved to be a useful cue that guides the reflection removal process [27, 31, 14], we develop the Concurrent Reflection Removal Network (CRRN) with a multi-task learning strategy, which concurrently estimates **B** and **R** under the guidance of $\nabla \mathbf{B}$. CRRN can be trained using multiple loss functions based on the ground truth of **B**, **R**, and $\nabla \mathbf{B}$, as shown in Figure 1. Given the input image **I**, we denote the dense prediction of **B**, **R** and $\nabla \mathbf{B}$ as follows:

$$(\mathbf{B}^\star, \mathbf{R}^\star, \nabla \mathbf{B}^\star) = \mathcal{F}(\mathbf{I}, \theta), \qquad (2)$$

where $\mathcal{F}$ is the network to be trained with $\theta$ consisting of all CNN parameters to be learned, and $\mathbf{B}^\star, \mathbf{R}^\star, \nabla \mathbf{B}^\star$ are the estimated values corresponding to their ground truth **B**, **R**, $\nabla \mathbf{B}$.

CRRN is implemented by designing two cooperative sub-networks. Different from the conventional two-stage framework, we combine the gradient inference and the image inference into one unified mechanism to do the two parts concurrently. For the gradient inference network (GiN), the input is a 4-channel tensor, which is the combination of the input mixture image and its corresponding gradients; it estimates $\nabla \mathbf{B}$ to extract the image gradient information from multiple scales and guide the whole image reconstruction process. The image inference network (IiN), takes the mixture image as the input and extracts background feature representations which describe the global structures and the high-level semantic information to estimate **B** and **R**. To allow the multiple estimation tasks to leverage information from each other, IiN shares the convolutional layers from GiN. The detailed architecture of GiN and IiN is introduced as follows:

**Gradient inference Network (GiN):** GiN is designed to learn a mapping from **I** to $\nabla \mathbf{B}$. As shown in Figure 1, the structure of GiN is a mirror-link framework with the encoder-decoder CNN architecture. The encoder part consists of five convolutional layers with stride equal to $1$ and five convolutional layers with stride equal to $2$. Each layer with stride $1$ is followed by the layer with stride $2$, which can progressively extract and down-sample features. In the decoder part, the features are upsampled and combined to reconstruct the output gradient without the reflection interference. In order to preserve the sharp details and avoid losing gradient information, the early encoder features are linked to its corresponding decoder layers with the same spatial resolution. An example result is shown in Figure 3, which demonstrates GiN successfully removes the gradients from reflection and remains the gradient belonging to the background.

**Image inference Network (IiN):** IiN is a multi-task learning network constructed on the basis of VGG16 network [22]. Recent works show that VGG16 network trained with large amount of data on high-level computer vision tasks can be well generalized to inverse imaging tasks such as shadow removal [19] and saliency detection [18]. To make the feature representations from the pre-trained VGG16 model suitable for our problem, we first replace the fully-connected layers in VGG16 model by a $3 \times 3$ convolutional layer [19] and then fine tune them for the reflection removal task.

After feature extractions with VGG16 net, we design a joint filtering network to predict **B** with multi-context features. It consists of two feature extraction layers and five transposed convolutional layers. We adopt the 'Reduction-A/B layers' from Inception-ResNet-v2 [24] as the 'Feature extraction layers A/B' in CRRN. Such a model is able to extract the scale invariant features by using multi-size kernels [11], but it is seldom used in image-to-image problems due to its decimated features caused by pooling layers. To make it fit our problem, we make two modifications: First, the pooling layers in the original model are replaced by two convolutional layers with $1 \times 1$ and $7 \times 7$ filter sizes, respectively; second, the stride of all convolutions are decreased to 1. The transposed convolutional layers in this part have a parallel framework which are composed of three sub-layers, as shown in Figure 1. We also adopt the residual network to help learn the mapping due to the narrow intensity range of the residual $(\mathbf{I} - \mathbf{B})$ [8].

**Multi-scale guided inference.** Multi-scale representations have shown to be effective in the extraction of image details for reflection removal [27] and other inverse imaging problems [12, 10]. To make full use of the multi-scale information of the decoder part in GiN, the output of each transposed convolutional layers of GiN is concatenated with the output of transposed convolutional layers in IiN at the same level, which is illustrated in Figure 1.

### 4.2. Loss function

Previous methods mainly adopt the pixel-wise loss function [7]. It is simple to calculate, but produces blurry predictions due to its inconsistency with human visual perception for natural images. To provide more visually pleasing results, we take the human perception into considerations when design our loss function.

In IiN, we adopt the perceptually motivated Structural similarity index (SSIM) [29] to measure the similarity between the estimated $\mathbf{B}^\star$ and $\mathbf{R}^\star$ and their corresponding ground truth. SSIM is defined as

$$\mathbf{SSIM}(x, x^\star) = \frac{(2\mu_x \mu_{x^\star} + C_1)(2\sigma_{xx^\star} + C_2)}{(\mu_x^2 + \mu_{x^\star}^2 + C_1)(\sigma_x^2 + \sigma_{x^\star}^2 + C_2)}, \quad (3)$$

where $\mu_x$ and $\mu_x^\star$ are the means of $x$ and $x^\star$, $\sigma_x$ and $\sigma_{x^\star}$ are the variances of $x$ and $x^\star$, and $\sigma_{xx^\star}$ is their corresponding covariances. SSIM measures the similarity between two images from the luminance, the contrast, and the structure. To make the values compatible with the common settings of the loss function in deep learning, we define our loss function for IiN as

$$\mathcal{L}^{\mathbf{SSIM}}(x, x^\star) = 1 - \mathbf{SSIM}(x, x^\star), \quad (4)$$

so that we can minimize it as that in the pixel-wise loss functions.

Despite its perceptual contribution, SSIM may cause changes of brightness and shifts of colors which makes the final results become dull [34], due to its insensitiveness to uniform bias. To solve this problem, we also introduce the $\mathcal{L}_1$ loss for the background layer to better balance brightness and color.

In GiN, the luminance and contrast components in SSIM become undefined. We therefore omit the dependence of contrast and luminance in the original SSIM and define the loss function for GiN as

$$\mathcal{L}^{\mathbf{SI}}(x, x^\star) = 1 - \mathbf{SI}(x, x^\star). \quad (5)$$

SI is used to measure the structural similarity between two images as demonstrated in [26], which is defined as

$$\mathbf{SI} = \frac{2\sigma_{xx^\star} + c}{\sigma_x^2 + \sigma_{x^\star}^2 + c}, \quad (6)$$

where all parameters share similar definitions as Equation (4).

Combining the above terms, our complete loss function becomes

$$\begin{aligned}\mathcal{L} =& \gamma \mathcal{L}^{\mathbf{SSIM}}(\mathbf{B}, \mathbf{B}^\star) + \mathcal{L}_1(\mathbf{B}, \mathbf{B}^\star) + \\ & \mathcal{L}^{\mathbf{SSIM}}(\mathbf{R}, \mathbf{R}^\star) + \mathcal{L}^{\mathbf{SI}}(\nabla \mathbf{B}, \nabla \mathbf{B}^\star),\end{aligned} \quad (7)$$

where the weighting coefficient $\gamma$ is empirically set as 0.8 in our experiments.

### 4.3. Training strategy

We have implemented CRRN using PyTorch[1]. To prevent overfitting, our network employs the multi-stage training strategy: GiN is first trained independently for 40 epochs with learning rate $10^{-4}$, then it is connected with IiN, and the entire network is fine-tuned end-to-end, which grants the two sub-networks more opportunities to cooperate accordingly. The learning rate for the whole network training is initially set to $10^{-4}$ for the first 30 epochs and the decreases to $10^{-5}$ for the next 20 epochs.

Prior works that use deep learning to solve the inverse imaging problems [33, 5] or layer separation problems [32] mainly optimize the whole network on patches with resolution $n \times n$ cropped from the whole images. However, many real-world reflections only occupy some regions in an image like the regional 'noise' [26], we call it regional properties of reflections. Training with the patches without obvious reflections could potentially degrade the final performance. To get avoid of such negative effects, CRRN is trained using whole images with different sizes. We adopt a multi-size training strategy by feeding images of two sizes: coarse scales $96 \times 160$ and fine scale $224 \times 288$, to make the network scale-invariant.

## 5. Experiments

To evaluate the performance of CRRN, we first compare with state-of-the-art reflection removal algorithms for both quantitative benchmark scores and visual qualities on the SIR$^2$ dataset [26]. We then conduct a self-comparison experiment to justify the necessity of the key components in CRRN. The SIR$^2$ dataset contains image triplets from controlled indoor setup and wild data. The indoor data are mainly designed to explore the influence of different parameters [26]. Since our method aims at removing reflections appeared in the wild scenes, we only evaluate on their wild dataset.

---

[1] http://pytorch.org/

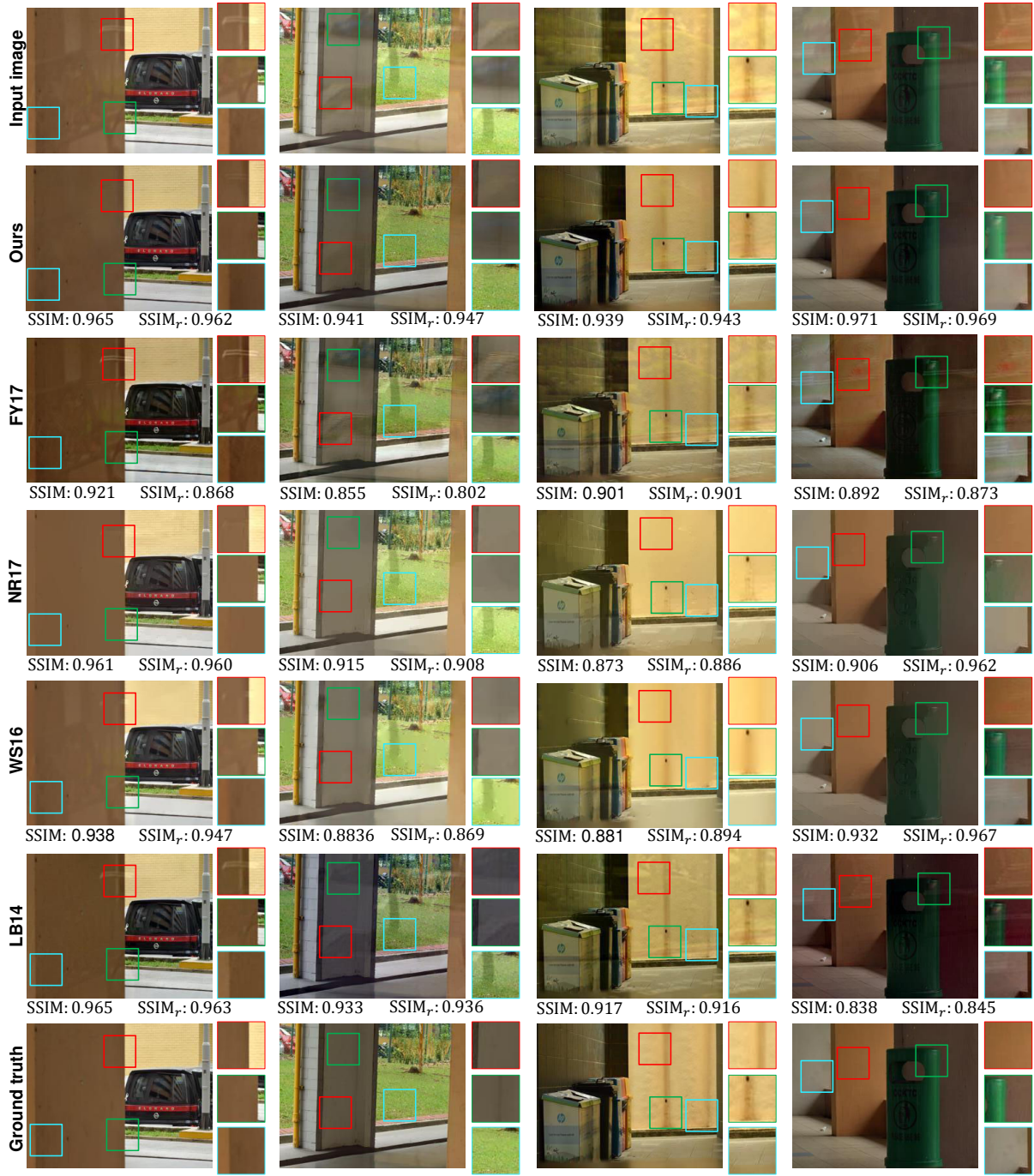

Figure 4. Examples of reflection removal results on four wild scenes, compared with FY17[7], NR17 [1], WS16 [27], and LB14 [16]. Corresponding close-up views are shown next to the images (with patch brightness ×2 for better visualization), and SSIM and SSIM$_r$ values are displayed below the images.

We adopt SSIM [29] and SI [26] as error metrics for our

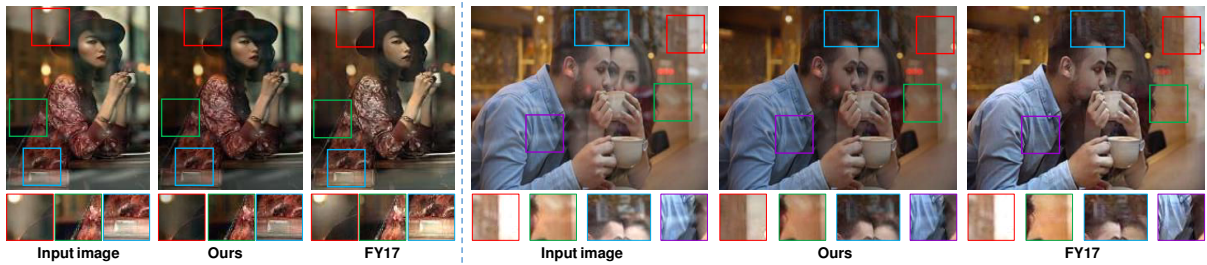
Figure 5. The generalization ability comparison with FY17 [7] on their released validation dataset.

Table 1. Quantitative evaluation results using four different error metrics, and compared with FY17[7], NR17 [1] WS16 [27] and LB14 [16].

|  | SSIM | SI | $\text{SSIM}_r$ | $\text{SI}_r$ |
|---|---|---|---|---|
| Ours | **0.895** | **0.925** | **0.861** | **0.890** |
| FY17 [7] | 0.867 | 0.902 | 0.812 | 0.847 |
| NR17 [1] | 0.884 | 0.903 | 0.850 | 0.880 |
| WS16 [27] | 0.876 | 0.910 | 0.843 | 0.881 |
| LB14 [16] | 0.833 | 0.920 | 0.801 | 0.861 |

quantitative evaluation, which are widely used by previous reflection removal methods [26, 31, 16]. Due to the regional properties of reflections, we experimentally observe that many existing reflection removal methods [1, 27, 16] may downgrade the quality of whole images, although they can remove the local reflections cleanly. The original definitions of SSIM and SI, which evaluate the similarity between $\mathbf{B}$ and $\mathbf{B}^\star$ in the whole image plane, may not reflect the performance of reflection removal unbiasedly. We therefore define the regional SSIM and SI, denoted as $\text{SSIM}_r$ and $\text{SI}_r$, to complement the limitations of global error metrics. We manually label the reflection dominant regions and evaluate the SSIM and SI values at these regions similar to the evaluation method proposed in [19, 25].

### 5.1. Comparison with the state-of-the-arts

We compare our method with state-of-the-art single-image reflection removal methods, including FY17 [7], NR17 [1], WS16 [27], and LB14 [16]. For a fair comparison, we use the codes provided by their authors and set the parameters as suggested in their original papers. For FY17 [7], we follow the same training protocol introduced in their paper to train their network using our training dataset.

**Quantitative comparison.** The quantitative evaluation results using four different error metrics and compared with four state-of-the-art methods are summarized in Table 1. The numbers displayed are the mean values over all 100 sets of wild images in the $\text{SIR}^2$ dataset. As shown in Table 1, CRNN consistently outperforms other methods for all four error metrics. The higher SSIM values indicate that our method recovers the whole background image with better quality, whose global appearance is closer to the ground truth. The higher SI values indicate that our method preserves the structural information more accurately. The higher $\text{SSIM}_r$ and $\text{SI}_r$ values mean that our method can remove strong reflections more efficiently in the regions overlaid with reflections than other methods. NR17 [1] shows the second best average performance with all error metrics.

**Visual quality comparison.** We then show examples of estimated background images by our method and four state-of-the-art methods in Figure 4 to check their visual quality. In these examples, our method removes reflections more effectively and recovers the details of the background images more clearly. All the non-learning based methods (NR17 [1], WS16 [27], and LB14 [16]) remove the reflections to some extent, but residual edges remain visible for the reflections that are not out of focus, and they also show some over-smooth artifacts when they are not able to differentiate the background and reflection clearly (*e.g.*, the result generated by WS16 [27] in the second column). LB14 [16] causes some color change in the estimated result (*e.g.*, the fourth column) partly due to the insensitivity of the Laplacian data fidelity term to the spatial shift of the pixel values [1]. NR17 [1] and LB14 [16] sometimes achieve similarly good quantitative values in SSIM (*e.g.*, the first column), but their estimated results still show obviously visible residual edges (the red box of LB14 [16] in the first column). The deep learning based method FY17 [7] is also good at preserving the image details and it does not cause the over-smooth artifacts as non-learning based method. However, the network in FY17 [7] is less effective in cleaning the residual edges comparing to CRRN. The SSIM and $\text{SSIM}_r$ values below each image also prove the advantage of our method.

**Comparing generality with FY17 [7].** The applicability to general unseen data of deep learning based methods is important yet challenging. To show the generalization ability of our method, we show results using released validation dataset from the project website of FY17 [7][2]. In this experiment, CRRN is still trained with our dataset described in Section 3.2 and strategy in Section 4.3, but for FY17 [7] we use the model released in their website (trained with their own data). Due to the lack of ground truth, only the visual quality is compared here. From the result shown in Figure 5, it is not surprised that FY17 [7] performs well using their trained model on their validation dataset, but CRRN also achieves reasonably good results and performs even better in some regions (*e.g.*, the red box in the left part of Figure 5). Recall that when FY17 [7] is trained with our data and tested on the SIR$^2$ dataset, its quantitative and qualitative performances are below our method as shown in previous experiments.

---
[2] https://github.com/fqnchina/CEILNet

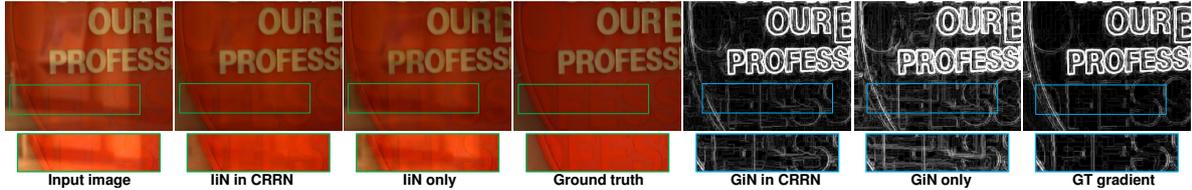

Figure 6. The output of IiN and GiN in CRRN against IiN and GiN only. Corresponding close-up views are shown below the images (with patch brightness ×1.6 for better visualization).

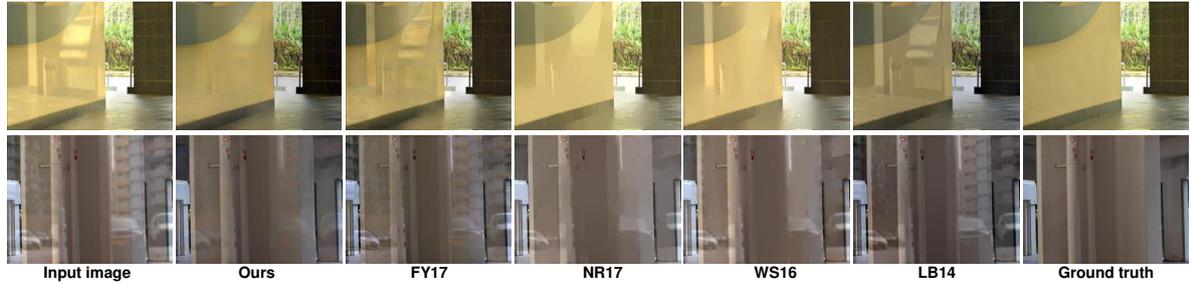

Figure 7. Extreme examples with whole-image-dominant reflections, compared with FY17 [7], NR17 [1] WS16 [27] and LB14 [16].

Table 2. Result comparisons of the proposed CRRN against CRRN using $\mathcal{L}_1$ loss in Equation (7) only and its sub-networks.

|  | SSIM | SI | $SSIM_r$ | $SI_r$ |
|---|---|---|---|---|
| IiN in CRRN | **0.895** | **0.925** | **0.861** | **0.890** |
| IiN in CRRN ($\mathcal{L}_1$) | 0.883 | 0.910 | 0.849 | 0.865 |
| IiN only | 0.867 | 0.892 | 0.843 | 0.859 |

### 5.2. Network analysis

CRRN consists of two sub-networks, *i.e.*, GiN and IiN. To further analyze the contribution of GiN and the perceptually motivated losses, we have trained three variant networks, one using $\mathcal{L}_1$ loss only, one using IiN only without the gradient feature layers and the other one using GiN only.

Table 2 shows the values using four error metrics of the two variant networks and the complete CRRN model. The comparisons between the results obtained by GiN in CRRN and GiN alone are shown in Figure 6. We can see that none of the three models perform better than the concurrent model using the perceptually motivated losses. When only using the pixel-wise losses, the performance of CRRN become worse. When removing GiN, IiN alone has relatively poor performance and the SSIM errors on the global and regional scales decreased to 0.867, compared with 0.895 by the concurrent model. From Figure 6, GiN in the CRRN model also outperforms GiN alone. The output of IiN only and GiN only remains more visible residual edges than that in CRRN as shown in the green and blue boxes in Figure 6. This demonstrates the effectiveness of the embedding mechanism in our network, where the two sub-network benefits each other in the whole estimation process.

### 6. Conclusion

We present a concurrent deep learning based framework to effectively remove reflection from a single image. Unlike the conventional pipeline that regards the gradient inference and image inference as two separate processes, our network unifies them as a concurrent framework, which integrates high-level image appearance information and multi-scale low-level features. Thanks to the newly collected real-world reflection image dataset and the corresponding training strategy, our method shows better performance than state-of-the-art methods for both the quantitative values and visual qualities and it is verified to be effectively generalized to other unseen data.

**Limitations.** The performance of CRRN may drop when the whole images are dominated by reflections. We show two examples on such extreme cases in Figure 7. In these examples, CRRN cannot remove the reflection completely and the estimated background still remains visible residual edges. However, even in this challenging examples, CRRN still removes the majority of reflections and restores the background details, which performs better than all other state-of-the-art methods. On the other hand, training a deep learning network directly on the images may suffer from gradient vanishing problem and the CNN may also introduce the color shift to the estimated image [8]. In the future, we will continue working on these parts to improve the generalization ability for dealing with challenging scenes.